\documentclass[10pt,twocolumn]{article}

\usepackage{multirow}
\usepackage{graphicx}
\usepackage{amsmath,amssymb}
\usepackage{times}
\usepackage{epsfig}
\usepackage{booktabs}
\usepackage{url}
\usepackage{hyperref}
\usepackage{subcaption}
\usepackage{wrapfig}
\usepackage{caption}
\usepackage{sidecap}
\sidecaptionvpos{figure}{c}
\def\sidecaptionrelwidth{0.32}
\usepackage{geometry}
\title{Auto3R: Automated 3D Reconstruction and Scanning via Data-driven Uncertainty Quantification}

\author{
Chentao Shen\textsuperscript{1,2} \and
Sizhe Zheng\textsuperscript{2} \and
Bingqian Wu\textsuperscript{2} \and
Yaohua Feng\textsuperscript{1} \and
Yuanchen Fei\textsuperscript{2,3} \and
Mingyu Mei\textsuperscript{1} \and
Hanwen Jiang\textsuperscript{4} \and
Xiangru Huang\textsuperscript{2}
}

\begin{document}
\twocolumn[{%
\renewcommand\twocolumn[1][]{#1}%
% \vspace{-0.5cm}
\maketitle
\vspace{-0.5cm}
\begin{minipage}{\textwidth}
\centering
\textsuperscript{1} Zhejiang University \hspace{1cm}
\textsuperscript{2} Westlake University \hspace{1cm}
\textsuperscript{3} Hunan University  \hspace{1cm}
\textsuperscript{4} Adobe Research
\\
shenchentao@zju.edu.cn huangxiangru@westlake.edu.cn
\end{minipage}

\includegraphics[trim={0pt 0pt 0pt 0pt}, width=\linewidth]{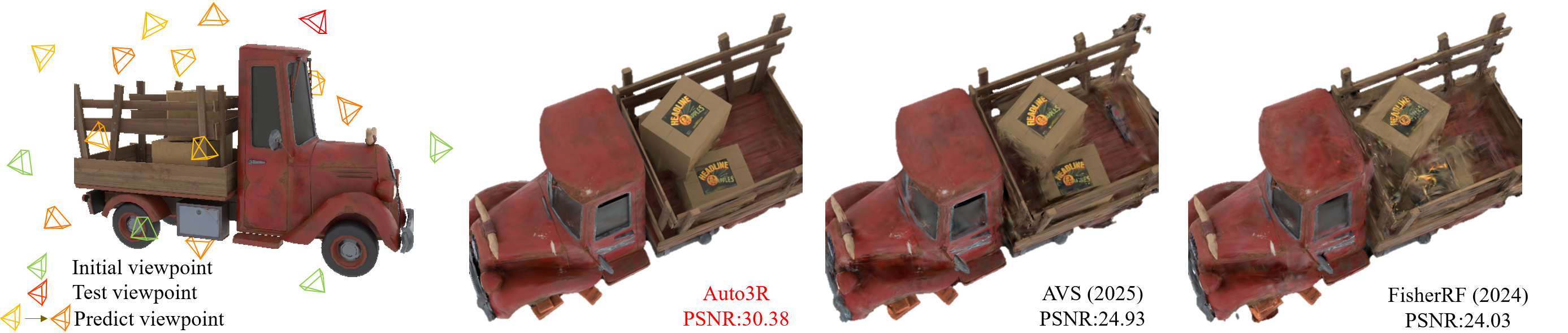}
\captionof{figure}{Left:  The scanning viewpoints selected by Auto3R, exhibiting a tendency to converge toward areas with occlusion. Right: Auto3R achieves accurate reconstruction quality and significantly outperforms the state-of-the-art methods. }
\vspace{0.4cm}
\label{fig:teaser}
}]

\begin{abstract}
Traditional high-quality 3D scanning and reconstruction typically relies on human labor to plan the scanning procedure. With the rapid development of embodied systems such as drones and robots, there is a growing demand of performing accurate 3D scanning and reconstruction in an fully automated manner. We introduce Auto3R, a data-driven uncertainty quantification model that is designed to automate the 3D scanning and reconstruction of scenes and objects, including objects with non-lambertian and specular materials. Specifically, in a process of iterative 3D reconstruction and scanning, Auto3R can make efficient and accurate prediction of uncertainty distribution over potential scanning viewpoints, without knowing the ground truth geometry and appearance. Through extensive experiments, Auto3R achieves superior performance that outperforms the state-of-the-art methods by a large margin. We also deploy Auto3R on a robot arm equipped with a camera and demonstrate that Auto3R can be used to effectively digitize real-world 3D objects and delivers ready-to-use and photorealistic digital assets. Our homepage: \url{https://tomatoma00.github.io/auto3r.github.io/}.
\end{abstract}    
\vspace{-0.1in}
\section{Introduction}
\label{sec:intro}

% Introduce active reconstruction
Acquiring high-fidelity 3D assets is essential for applications in gaming, film, and virtual reality. While recent advances in 3D reconstruction—particularly approaches based on 3D Gaussian Splatting (3DGS)~\cite{3dgs} and neural rendering~\cite{mildenhall2021nerf}, have greatly improved visual quality, the reconstruction pipeline itself remains costly and labor-intensive. In practice, generating a high-quality 3D model still requires capturing a large number of informative views, often involving manual trajectory planning and repeated quality inspection~\cite{chen2024gennbv,kriegel2012next,mostegel2016uav}.
To address this challenge, we study the problem of Active Reconstruction, which aims to automate and accelerate the data acquisition process. It is commonly formulated as an iterative loop: from the currently captured images, the system performs reconstruction, selects the next viewpoints expected to yield the greatest improvement, and then captures these views. This procedure reduces human involvement, improves efficiency, and is evaluated by the final reconstruction quality as well as the resources consumed (e.g., number of views or total runtime).

\begin{figure}
    \centering
    \includegraphics[width=0.95\linewidth]{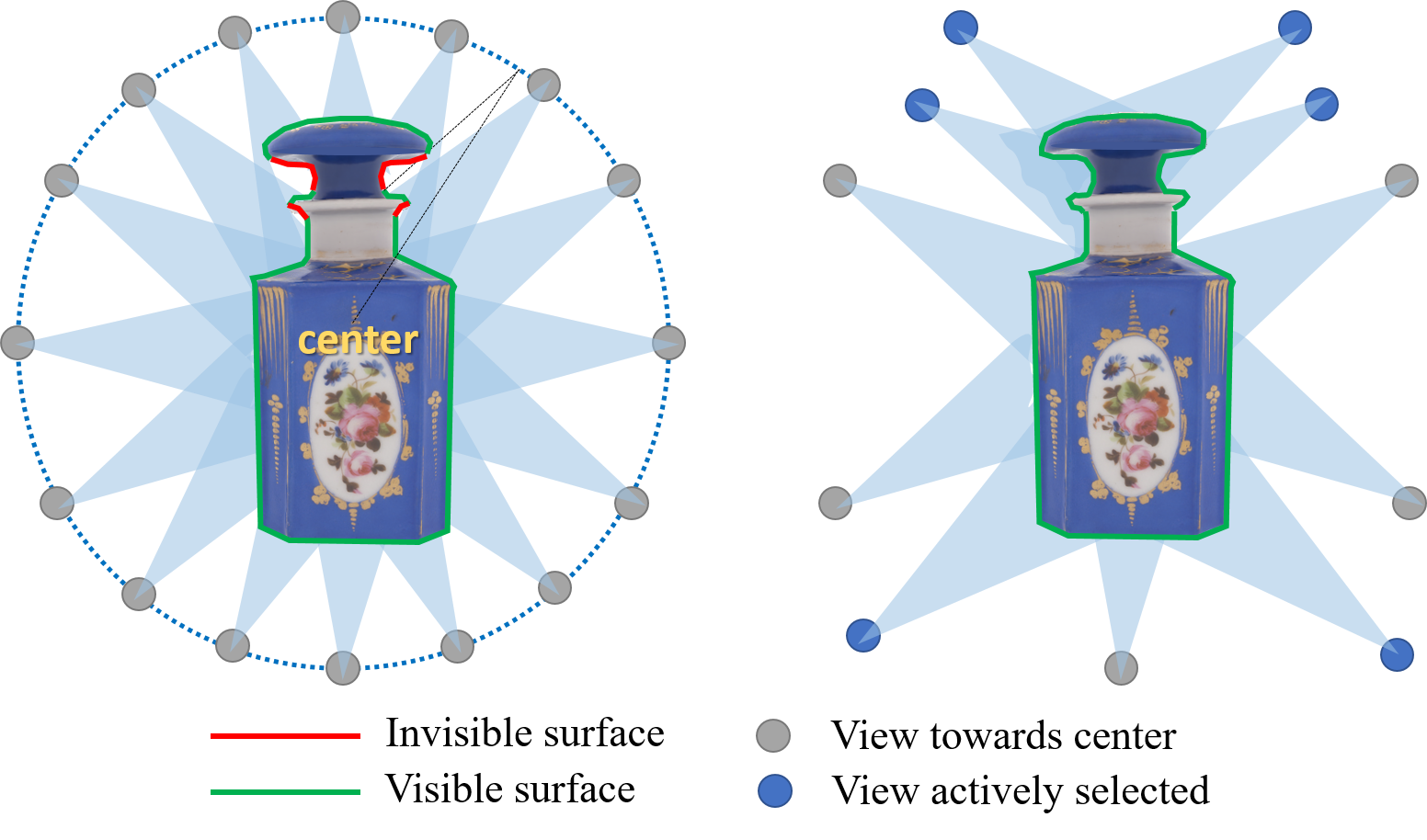}
    \vspace{-0.1in}
    \caption{Left: Reconstruction from randomly sampled viewpoints can lead to incomplete observation (highlighted with red surface) and thus low-quality results. Right: Active planning formulated with uncertainty prediction solves the problem by selectively chose observation viewpoints for scanning.}
    \label{fig:placeholder}
    \vspace{-0.15in}
\end{figure}

% The Key Challenges of AR and existing methods
The key challenge in active reconstruction lies in developing a reliable and efficient \textbf{Uncertainty Quantification (UQ)} model to guide viewpoint selection. This is non-trivial, as the UQ model must determine where to scan next based solely on the current reconstruction results. Ideally, it should accurately identify regions of low reconstruction quality while avoiding redundant views, thereby improving reconstruction efficiency and overall quality.

Traditional UQ models~\cite{fisherrf,xie2025gauss} rely on analytical metrics derived from statistics and information theory, such as Fisher information or mutual information. However, these metrics serve only as \textit{indirect proxies} of the true reconstruction error. While they approximate the expected information gain from adding a new view, they fail to capture complex photometric and geometric inconsistencies arising from material properties or occlusions. In addition, their reliance on \textit{analytical approximations} (e.g., local linearization or Hessian estimation) leads to significant computational overhead, hindering real-time performance.

Recent works such as PUN~\cite{zhang2025peering} and Active-View-Selector (AVS)~\cite{wang2025active} adopt data-driven approaches that learn uncertainty directly from rendered or captured images. However, these methods remain limited. PUN predicts a global uncertainty field from a single input image, disregarding the intermediate reconstruction states and thus lacking scene-specific adaptivity. AVS improves upon this by comparing rendered candidate views with existing captures, yet it only estimates 2D image-level reconstruction errors (e.g., SSIM) without explicitly reasoning about 3D geometry or depth reliability.

In contrast, our method introduces a \textbf{joint 2D–3D uncertainty formulation} that integrates rendered color, depth, and their uncertainties within a unified data-driven framework. Unlike PUN and AVS, Auto3R explicitly couples photometric and geometric uncertainty, enabling more precise identification of ambiguous or incomplete regions. This design allows the system to select viewpoints that maximize the expected reduction in reconstruction uncertainty while remaining computationally efficient.

To summarize, our contributions are:
\begin{itemize}
    \item \textbf{A fully automated 3D reconstruction framework.} We introduce \textbf{Auto3R}, the depth aware data-driven uncertainty quantification (UQ) model tailored for active 3D reconstruction and scanning.
    \item \textbf{Data-driven uncertainty quantification.} We design a dual-branch UQ model that fuses color and depth cues via depth-aware blending and reweighting for precise and efficient viewpoint selection.
    \item \textbf{Real-world robotic deployment.} We extend Auto3R with a video-based UQ module for path-level uncertainty estimation, enabling efficient and adaptive robotic scanning in real-world environments.
\end{itemize}
\section{Related Works}
\label{sec:related}

%\subsection{Active View Selection}

%(1) Search: Point cloud-based search \cite{border2018see,border2020,border2024see}, surface-based search \cite{kriegel2011,lee2020tro,wu2014tog}, voxel-based search \cite{krain2011icra,vasquez2014,vasquez2017,daudelin2017,delmerico2018,pan2022ral,pan2023cviu}, NeRF-based search \cite{lee2022ral,sunder2023icra}

%(2) Learning: reinforcement learning \cite{Peralta2020eccv,zeng2022icra}, voxel-based networks \cite{miguel2020prl,Vasquez2021mvap,pan2025tro,jia2025pbnbv}, point cloud based networks \cite{zeng2020iros,han2022icra}. new: ~\cite{pan2024icra}, voxel+3DGS-based ~\cite{jin2025ral},,neunbv \cite{neunbv}

Traditional reconstruction methods often depend on a fixed, manually captured set of images (e.g., spherical or grid-based sampling). Such approaches typically require a large number of input views to achieve satisfactory reconstruction quality, leading to inefficiencies in both data acquisition and computational cost. To address this limitation, automated reconstruction methods have been proposed to actively predict the next viewpoint that is expected to maximize the improvement in reconstruction quality. 
%Currently, mainstream scene representations are two types, differing in information storage and rendering, which directly shapes active reconstruction design. One is NeRF (Neural Radiance Fields)\cite{mildenhall2021nerf}, an implicit representation that maps 3D positions/viewing directions to color/volume density via MLP. It enables fine reconstruction but requires many calibrated viewpoints, with time-consuming rendering/optimization that fails real-time needs. The other is 3DGS (3D Gaussian Splatting)\cite{3dgs}, an explicit representation that models scenes with 3D Gaussian primitives (position, covariance, etc.). It achieves real-time rendering via efficient rasterization, better suited for fast viewpoint evaluation in active reconstruction.

% has developed differenet technical paths: The earliest approach is based on Image Quality Assessment (IQA), which avoids complex 3D computations by selecting the viewpoint with the ``worst 2D rendering quality'' in current reconstruction. \cite{golestaneh2022no, yang2022maniqa, ke2021musiq, talebi2018nima} use CNNs (for local features) and ViT~\cite{dosovitskiy2020image} (for non-local dependencies) to address traditional IQA’s problems. In order to better focus on the core area of 3D reconstruction, TOPIQ~\cite{chen2024topiq} also adds semantic guidance to correct this misjudgment. Though these methods are efficient and mature, their accuracy is lower than 3D-based schemes.
\noindent \textbf{Automated 3D Reconstruction} has been extensively studied, evolving from explicit geometric representations (e.g., point clouds, meshes, voxels) to neural implicit and hybrid representations (e.g., NeRF, SDF, 3DGS).
For \textit{explicit representations}, point cloud–based methods~\cite{border2018see,border2020,border2024see} identify under-observed regions via point density and visibility analysis, while mesh-based approaches~\cite{kriegel2011,lee2020tro,wu2014tog} detect incomplete surfaces from curvature and normal consistency. Voxel-based methods~\cite{krain2011icra,vasquez2014,vasquez2017,daudelin2017,delmerico2018,pan2022ral,pan2023cviu} estimate view utility through occupancy and uncertainty measures. However, these geometry-driven techniques rely on handcrafted priors, making them sensitive to noise and less generalizable to complex or reflective surfaces.
With \textit{NeRF-based active reconstruction}, models such as ActiveNeRF~\cite{uqnerfeccv23}, NeRFDirector~\cite{xiao2024nerf}, and NVF~\cite{xue2024neural} estimate information gain or visibility to guide view selection, but remain computationally intensive and limited to synthetic settings.
Recently, \textit{3DGS-based methods} have shown promise for efficient active reconstruction. ActiveGS~\cite{jin2025ral}, ActiveSplat~\cite{li2025activesplat}, FisherRF~\cite{fisherrf}, and GauSS-MI~\cite{xie2025gauss} adopt uncertainty or visibility-driven selection, yet still rely on analytic approximations (e.g., Hessian or mutual information) that are costly and weakly correlated with perceptual quality.
In contrast, our \textit{Auto3R} introduces a data-driven uncertainty quantification model that learns directly from rendered color and depth, achieving accurate, material-robust, and efficient view planning.

\begin{figure*}
    \centering
    \includegraphics[width=0.9\linewidth]{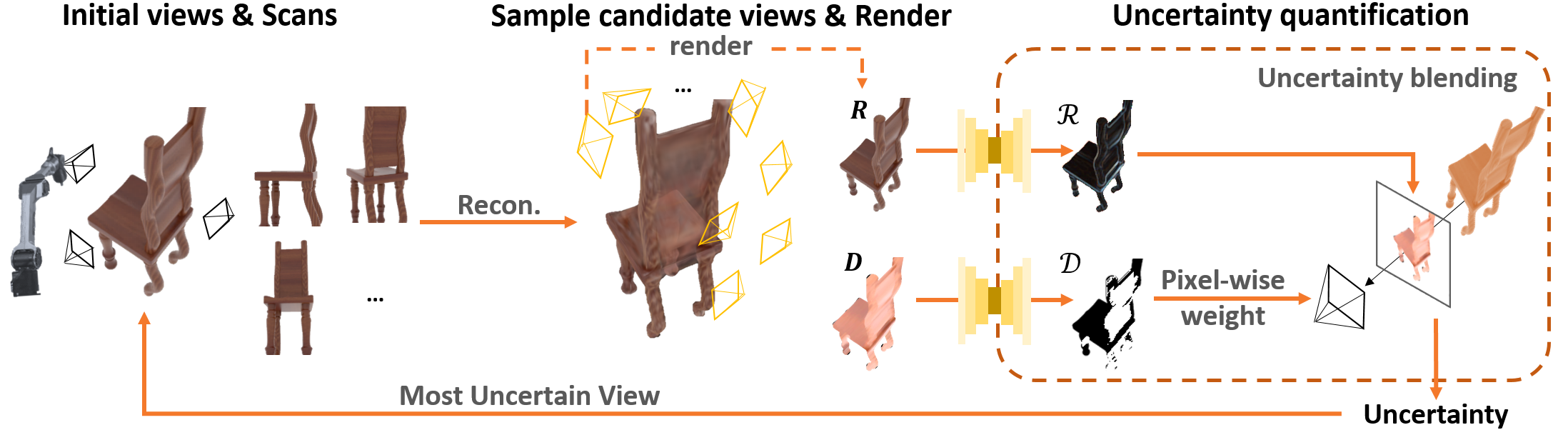}
    \caption{\textbf{An illustration of our automated scanning and reconstruction methods.} Given some images from scanning views, we reconstruction them then render the image and depth map on some candidate scanning viewpoints. We proposed an uncertainty quantification on them, obtaining the uncertainty of each candidate viewpoint. Finally scan on the most uncertain viewpoints, and repeat the process.}
    \vspace{-0.1in}
    \label{fig:3pipline}
\end{figure*}

\noindent \textbf{Uncertainty Quantification.}
The core of automated reconstruction lies in accurately estimating the \textit{uncertainty} of the current reconstruction. Uncertainty quantification (UQ) is central to active reconstruction, as it guides the system to prioritize viewpoints that most effectively reduce geometric or photometric ambiguity while avoiding redundant captures.
Existing methods can be categorized by the \textit{space in which uncertainty is defined}.
Methods that operate in the \textit{3D domain} (hereafter \textit{3D uncertainty}) typically attach uncertainty parameters to geometric primitives such as voxels or Gaussians. Early works estimate uncertainty from visibility and occupancy~\cite{kriegel2011,krain2011icra}, while later studies measure geometric confidence through distribution variance~\cite{uqtnnls25,uq3dgsnips24,neurar,uq3dv21,uqoccral23}. Information-theoretic models~\cite{PUP3DGS,fisherrf,jin2025ral,xie2025gauss} evaluate potential information gain using Fisher or Shannon information, and recent works~\cite{uqwacv24} model uncertainty in latent feature space. Although effective, these 3D-based approaches rely on analytical approximations or handcrafted priors, often missing photometric inconsistencies or reflectance-induced ambiguities.
Other methods define uncertainty in the \textit{2D rendering space}. Rendering-based methods~\cite{neunbv,sunder2023icra} estimate pixel-wise variance, whereas noise-based approaches~\cite{uqnormal,wang2025active,uqnerfeccv23} infer uncertainty from observation noise using fixed priors. NeRF-based UQ models~\cite{uqnerfcvpr24,uqsa24,xiao2024nerf} extend this idea to per-ray estimation but remain computationally expensive.
In contrast, our approach jointly learns 2D and 3D uncertainty representations directly from rendered color and depth, bridging appearance and geometry. This data-driven formulation captures both photometric artifacts and structural incompleteness, enabling accurate, efficient, and generalizable UQ for active reconstruction.

\section{Overview}
\label{sec:overview}
% \sct{Chapter 3 has modified first rounds.}

In this section, we briefly introduce the key components of our automated reconstruction algorithm.
% \begin{figure*}
%     \centering
%     \includegraphics[width=0.95\linewidth]{figures/3pipmain.png}
%     \caption{An illustration of our automated scanning and reconstruction methods. Given some images from scanning views, we reconstruction them then render the image and depth map on some candidate scanning viewpoints. We proposed an uncertainty quantification on them, obtaining the uncertainty of each candidate viewpoint. Finally scan on the most uncertain viewpoints, and repeat the process.}
%     \label{fig:3pipline}
% \end{figure*}

\textbf{Framework.} As shown in Figure~\ref{fig:3pipline}, we begin reconstruction from a small set of initial viewpoints. After several iterations, a set of candidate viewpoints is sampled and evaluated based on the uncertainty of the current reconstruction under each candidate view. The candidate viewpoint with the highest uncertainty is then selected for scanning, and the newly captured image is added to the training set for iterative refinement

\textbf{3DGS-based Reconstruction.} We adopt 3D Gaussian Splatting (3DGS) as our scene representation due to its fast convergence and suitability. For viewpoint selection, we render both color and depth map from the 3D Gaussians for each candidate viewpoint, enabling uncertainty evaluation for view planning.

\textbf{Data-driven Uncertainty Prior.} We propose a module to quantify the uncertainty of rendered color and depth maps, where uncertainty typically manifests as Gaussian-induced blur, aliasing artifacts, or structural distortions. A lightweight image network takes the rendered images and depth maps as input and predicts per-view \textit{render uncertainty} $\mathcal{R}$ and \textit{depth uncertainty} $\mathcal{D}$ for each candidate viewpoint.

\textbf{Uncertainty Quantification (UQ).}
Our method jointly models uncertainty inferred from both 2D appearance and 3D geometry. We introduce a depth-aware uncertainty blending scheme that combines the render uncertainty $\mathcal{R}$ with depth information using depth values as blending weights. Since depth itself carries uncertainty, each depth layer is further re-weighted by its corresponding depth uncertainty before the final aggregation.

\textbf{Automated Scanning.}
We further extend our UQ model to plan a continuous scanning path consisting of multiple viewpoints, rather than selecting a single view at a time. A lightweight video-based network takes as input a sequence of rendered images along the candidate path and predicts an overall uncertainty score, which guides the robot in selecting the optimal scanning trajectory.

\section{Method}
\label{sec:method}
In this section, we introduce the detail of data-driven image uncertainty prior (Sec.~\ref{sec:4.1}), uncertainty quantification (Sec.~\ref{sec:4.2}), and the extension to real-world (Sec.~\ref{sec:4.3}).

\subsection{Data-driven Image Uncertainty Map}
\label{sec:4.1}

\begin{figure}
\centering
\includegraphics[width=\linewidth]{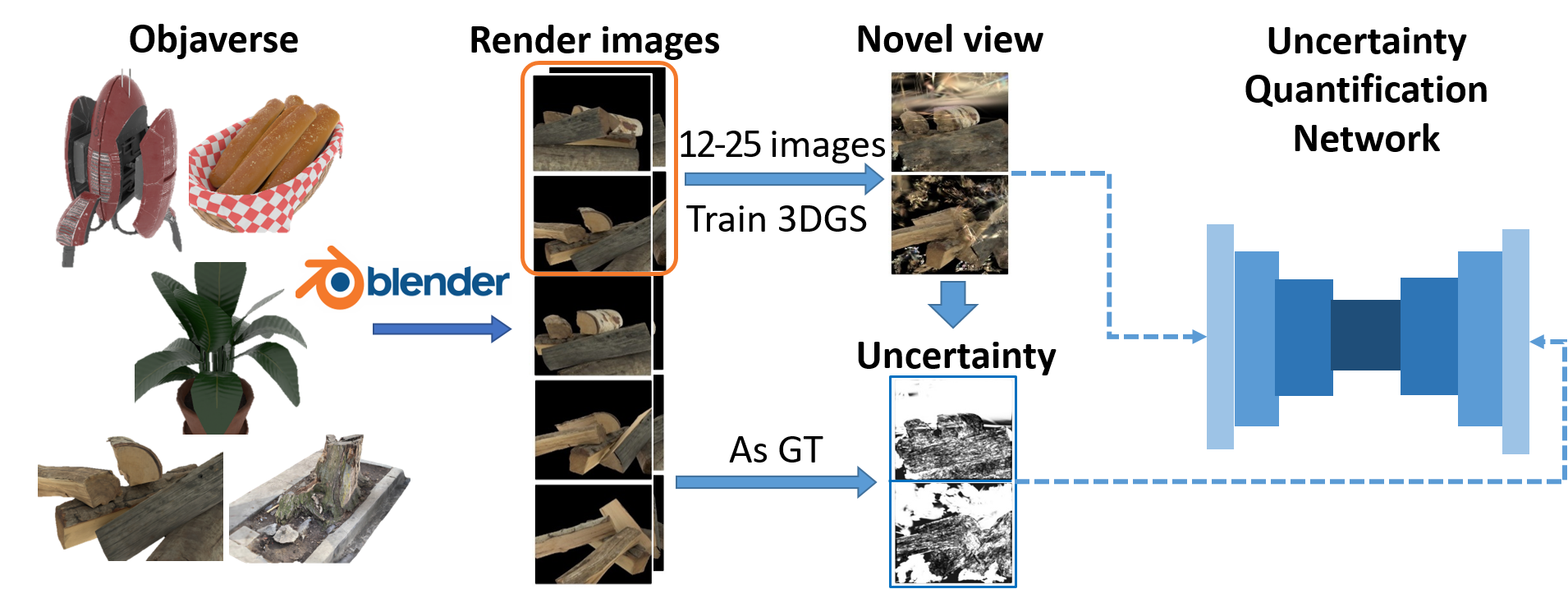}
\caption{\textbf{Training pipeline of our data-driven uncertainty map network.} We render 3000 Objaverse objects with random illumination, train 3DGS using 12–25 sparse views, and render novel views for training the model. The uncertainty is then formulated as the SSIM between the rendered and ground-truth novel views.}
\vspace{-0.05in}
\label{fig:trainpipe}
\end{figure}

As described in Sec.~\ref{sec:overview}, we represent each object using 3D Gaussians, denoted as $\mathcal{G}$. Each Gaussian $\mathcal{G}^i$ has a mean $\boldsymbol{\mu}^i=(x^i,y^i,z^i)$, covariance $\boldsymbol{\Sigma}^i$, color $c^i$, and opacity $\alpha^i$, where the first two define spatial shape and the latter two define appearance.

When projected to a viewpoint, Gaussians are splatted and composited in depth order. The rendered color and depth for pixel $(u,v)$ are computed as:
\begin{equation}
\mathbf{R}(u,v)=\sum_{i} c^i \alpha^i \prod_{n=1}^{i-1} (1-\alpha^n),
\label{eq:3dgs-r}
\end{equation}
\begin{equation}
\mathbf{D}(u,v)=\sum_{i} z^i \alpha^i \prod_{n=1}^{i-1} (1-\alpha^n).
\label{eq:3dgs-d}
\end{equation}

Due to incomplete viewpoint coverage or local overfitting, artifacts such as color distortion, ghosting, and inconsistent geometry often appear in rendered images or depth maps. These visible errors directly indicate regions of high \textbf{reconstruction uncertainty}. We therefore model these spatially varying artifacts as a \textbf{uncertainty map}, which can be inferred from a single rendered image or depth map. Thus, we employ two lightweight ResNet-50–based image-to-image networks to predict pixel-wise uncertainty maps from rendered color and depth, respectively. Both networks are trained in a self-supervised manner, requiring no additional ground-truth uncertainty labels.

% Since these cues can be perceived from a single rendering without reference, we employ two lightweight ResNet-50–based image-to-image networks to predict pixel-wise uncertainty maps from rendered color and depth respectively. Both networks are trained in a self-supervised manner without additional ground-truth uncertainty labels.

The training process is illustrated in Figure~\ref{fig:trainpipe}. We select 3,000 objects from the MaterialAnything subset~\cite{huang2025materialanything} of Objaverse~\cite{deitke2023objaverse} and render them under random HDR environment lighting in Blender 3.2.2. For each object, 60 random camera poses are generated to ensure full object coverage. Among them, 12–25 views are used to train the 3DGS model, while the remaining 35–48 novel views are rendered for uncertainty supervision.
For scene-level experiments, we train our model on the map-free-reloc dataset~\cite{arnold2022mapfree} using the same protocol. Because the number of training views varies, the resulting reconstructions naturally exhibit a wide quality range -- from coarse to highly accurate -- providing diverse supervision for uncertainty learning. We use the pixel-wise SSIM between rendered and ground-truth images as the training signal, enabling the model to learn the mapping from local visual artifacts to uncertainty.

\subsection{Uncertainty Quantification}
\label{sec:4.2}

\begin{figure}
\centering
\includegraphics[width=0.9\linewidth]{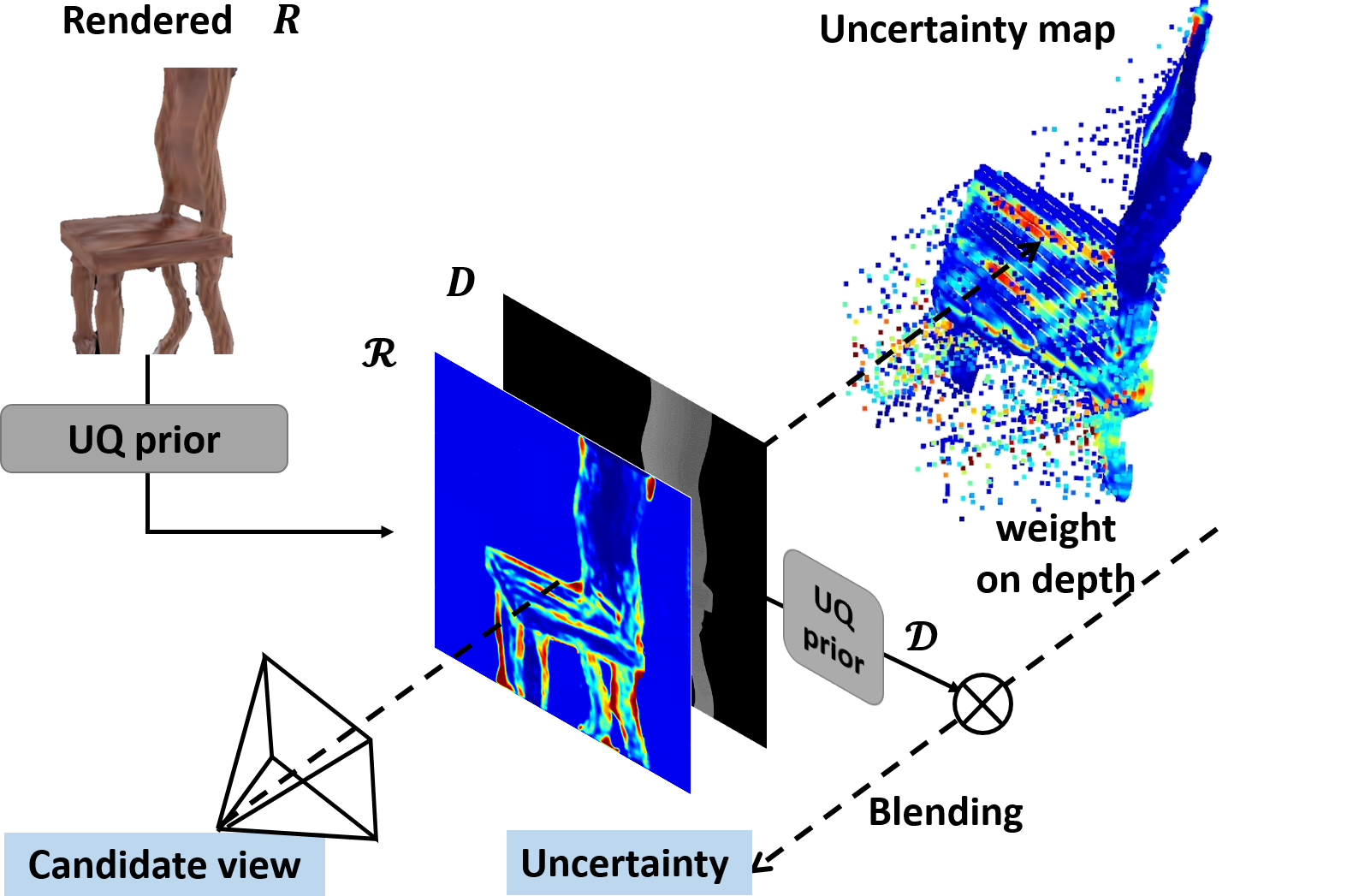}
\caption{\textbf{Illustration of our uncertainty quantification.} Rendered-image uncertainty is integrated along depth using perspective projection, where deeper pixels receive higher weights, and is further reweighted by depth-map uncertainty.}
\vspace{-0.1in}
\label{fig:uquq}
\end{figure}

After training the UQ network described in Sec.~\ref{sec:4.1}, given a Gaussian-rendered RGB image $\mathbf{R}$ and its corresponding depth map $\mathbf{D}$, we obtain two predicted uncertainty maps: the image-based uncertainty $\mathcal{R}$ and the depth-based uncertainty $\mathcal{D}$.
We note that relying on either modality alone provides an incomplete estimation, as image-based UQ is sensitive to texture and lighting variations, whereas depth-based UQ is less robust to photometric artifacts. Thus, we integrate both cues into a unified uncertainty quantification framework that jointly leverages appearance and geometry.

\begin{figure*}
    \centering
    \includegraphics[width=0.96\linewidth]{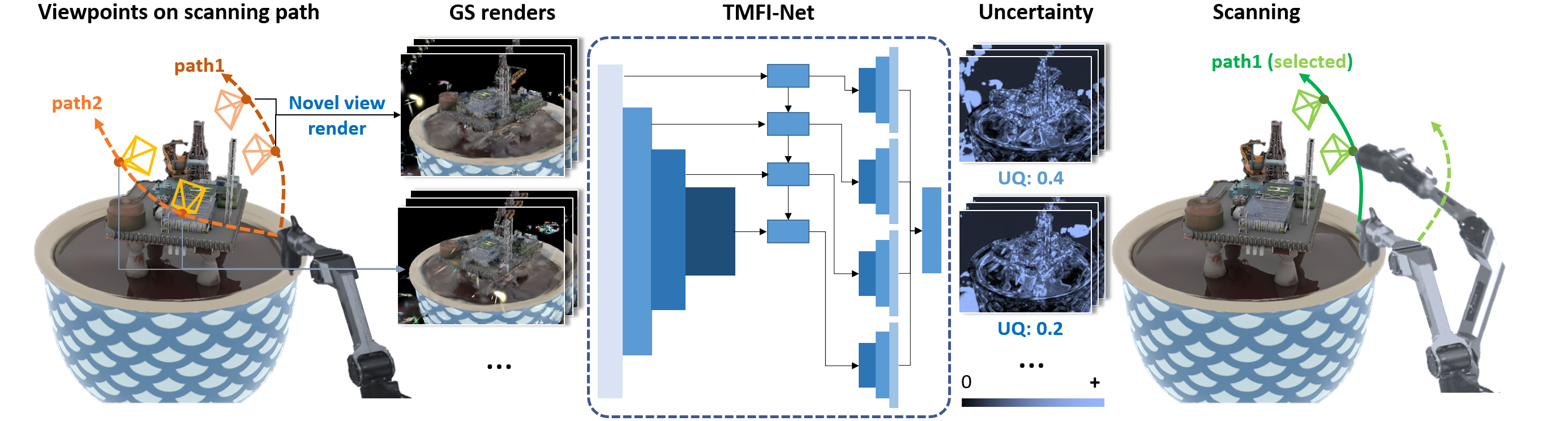}
    \caption{\textbf{The extension of uncertainty quantification for real-world scanning.} We measure uncertainty on rendered image sampled along potential active trajectories. Then we select the trajectory with larger uncertainty to scan.}
    \label{fig:videouq}
    \vspace{-0.1in}
\end{figure*}

\textbf{Depth-aware Blending.}
For perspective projection, the apparent size of a 3D Gaussian on the image plane is inversely proportional to its distance from the camera: a farther Gaussian projects to a larger area and influences more pixels. Consequently, the number of Gaussians contributing to a pixel increases approximately with the square of its depth.
To account for this geometric relationship, we propose a \textbf{depth-aware blending} strategy that assigns higher weights to deeper pixels when aggregating uncertainty across the image. The blending is formulated as:
\begin{equation}
\label{eq:blend}
l_{\text{blend}} = \sum_{(u,v)} \mathbf{D}(u,v) \cdot \mathcal{R}(u,v),
\end{equation}
where $\mathbf{D}(u,v)$ denotes the depth value and $\mathcal{R}(u,v)$ the corresponding image-based uncertainty at pixel $(u,v)$.
This operation effectively emphasizes regions with higher projection overlap and reconstruction ambiguity.

\textbf{Depth-Uncertainty Reweighting.}
During the early stages of reconstruction, depth estimates may be unreliable due to sparse observations.
To reduce the impact of inaccurate depth, we reweight each pixel’s contribution according to its predicted depth uncertainty. Pixels with higher confidence (i.e., lower $\mathcal{D}$) receive larger weights, leading to the following formulation:
\begin{equation}
\label{eq:blend2}
l_{\text{blend*}} = \sum_{(u,v)} \mathbf{D}(u,v) \cdot \mathcal{R}(u,v) \cdot (1-\mathcal{D}(u,v)).
\end{equation}
Finally, we integrate both blending terms with global regularization over the uncertainty maps:
\begin{equation}
\label{eq:totaluq}
l = l_{\text{blend}} + \lambda_0 l_{\text{blend*}} + \lambda_1 \sum_{(u,v)} \mathcal{R}(u,v) + \lambda_2 \sum_{(u,v)} \mathcal{D}(u,v),
\end{equation}
where $\lambda_0$, $\lambda_1$, and $\lambda_2$ control the relative weights of the reweighted blending term and the regularization of image- and depth-based uncertainties, respectively.
This formulation jointly models appearance and geometric uncertainty, yielding consistent predictions across varying levels of reconstruction completeness and scene complexity.

\subsection{Scanning Path UQ for Robotics Task}
\label{sec:4.3}

In real-world scenarios, the camera is typically mounted on a robotic platform that captures images continuously during motion. Capturing images along the trajectory introduces only minor efficiency overhead compared to pausing at discrete target poses, as the overall acquisition time primarily depends on the number of robot movements.

We extend our image-based UQ model to take a \textbf{sequence of images} as input, enabling the system to determine the optimal scanning path for sequential image capture.
We adapt the encoder of TMFI-Net~\cite{TMFI} and design an upsampling module to predict per-frame uncertainty, as illustrated in Figure~\ref{fig:videouq}. Compared with single-image prediction networks, this sequence-based model achieves higher efficiency and a more comprehensive understanding of object surfaces.

During training, each object is rendered with 60 views, and a 3DGS model is trained using a subset of them following Sec.~\ref{sec:4.1}. We then generate multiple paths connecting nearby viewpoints and interpolate intermediate poses along each path. Continuous image capture is simulated by rendering frames along these trajectories. As before, we use the pixel-wise SSIM between rendered and ground-truth images as the supervision signal for training TMFI-Net.
\section{Experiments}
\label{sec:exp}

\begin{figure}
    \centering
    \includegraphics[width=0.96\linewidth]{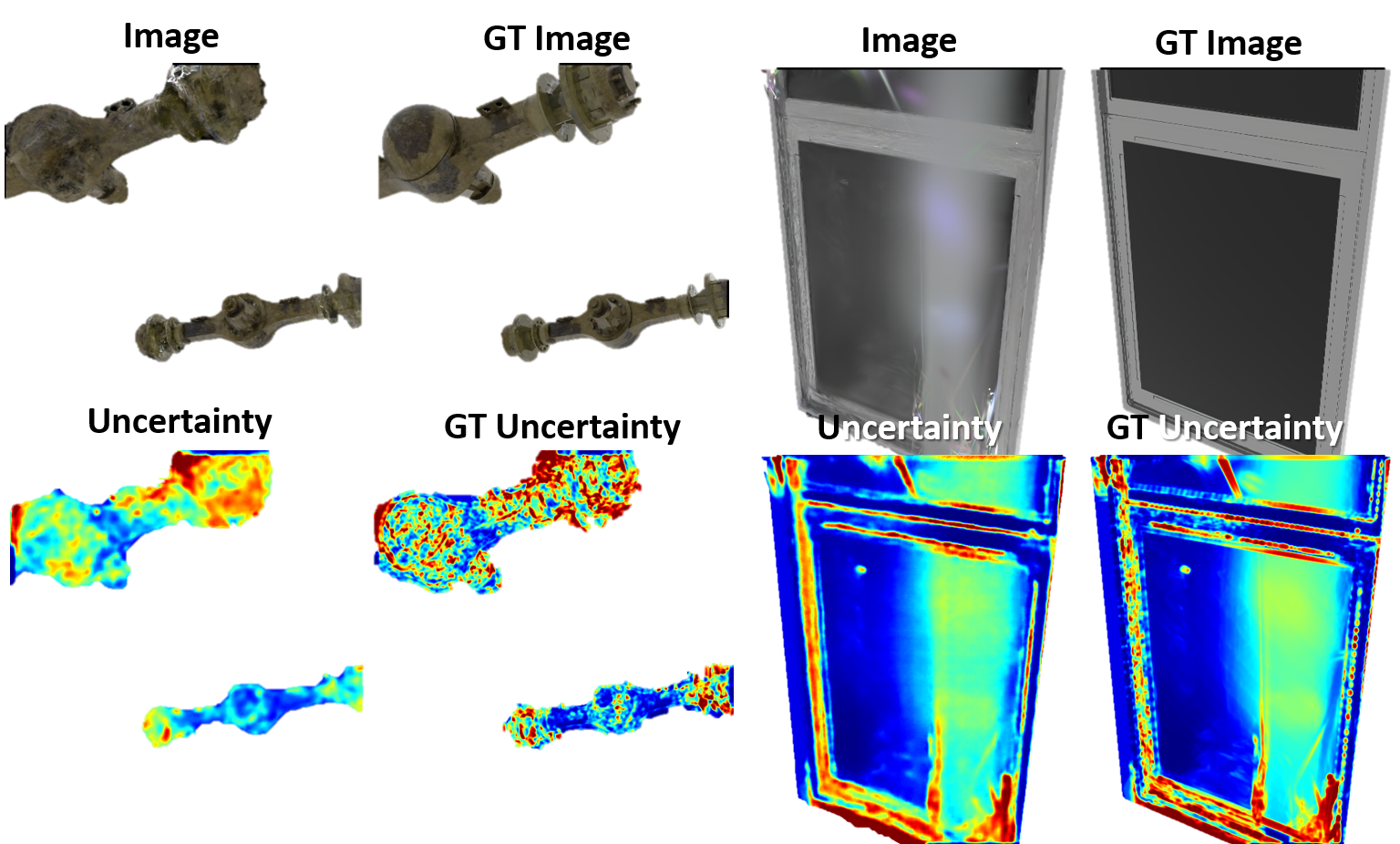}
    \vspace{-0.12in}
    \caption{\textbf{Visualization of Uncertainty Predictions.} Top row: rendered image (input) and ground truth (GT). Bottom row: predicted and GT uncertainty maps. Color encoding from blue (low) to red (high) indicates uncertainty magnitude.}
    \vspace{-0.1in}
    \label{fig:5uq}
\end{figure}

We test Auto3R on three different active reconstruction tasks, active normal object reconstruction, active specular object reconstruction and active scene reconstruction.
\subsection{Experimental Setup}
\textbf{Dataset}. For objects reconstruction, the normal and specular objects are selected from Objaverse~\cite{deitke2023objaverse}, containing objects covering various types such as animated objects, bound characters, etc. Notably, we provide 256 candidate viewpoints for view select, and 256 test viewpoints for evaluation, to simulate real-world scanning task, we \textbf{did not input initial 3D points before training}. For scene reconstruction, we use Mip-NeRF360~\cite{barron2022mip} dataset, which covers complex objects and contexts under different lighting conditions of indoors and outdoors scenes. 

\textbf{Baseline}. For fair comparison, all baselines are evaluated under the active view-selection protocol. The baselines can be grouped into 2 categories. The first is based on image-based uncertainty, including AVS~\cite{wang2025active}, we also adapt other image uncertainty/quality quantification methods TOPIQ~\cite{chen2024topiq}, TRES~\cite{golestaneh2022no}, MANIQA~\cite{yang2022maniqa}, MUSIQ~\cite{ke2021musiq} to the frame work of activate reconstruction. On the other hand, we also compare with baselines based on 3D uncertainty, such as FisherRF~\cite{fisherrf}and Gauss-MI~\cite{xie2025gauss}. As for AVS, We re-implemented the CNN version and trained it on our dataset under the same conditions for fair comparison.
% Finally, we compared our robotic reconstruction with  ActiveGS\cite{jin2025ral}, and Next best sense\cite{nextbestsense}
%We select ActiveNeRF\cite{uqnerfeccv23}, NVF\cite{xue2024neural} and NeRFDirector\cite{xiao2024nerf} for the comparison. 

\textbf{Scheduling}. For both scene reconstruction and objects reconstruction, we train the 3DGS model for 30000 iterations. Over these 30000 iterations, we add one new view at specific intervals following the schedule of FisherRF~\cite{fisherrf}: [400, 900, 1500, 2200, 3000, 3900, 4900, 6000, 7200, 8500, 9900, 11400, 13000, 14700, 16500, 18400]. In all methods, we provide a final set of 20 training views, including the initial views. 

\textbf{Metrics}. We evaluate reconstruction quality using standard image-based metrics that measure photometric fidelity and perceptual similarity. Specifically, we report PSNR, SSIM, and LPIPS for novel view synthesis under test viewpoints. For comprehensive evaluation, each reconstruction task includes both the average performance across all viewpoints and the worst 5\% performance to capture failure cases.

\begin{table}[t]
\centering
\footnotesize
\caption{\textbf{Quantitative results of active object reconstruction on the Objaverse dataset}. We report average and worst-5\% PSNR, SSIM, and LPIPS over all test views. Auto3R achieves the best overall reconstruction quality across all metrics.}
\vspace{-0.05in}
\label{tab:noramaltable}
\resizebox{\columnwidth}{!}{
\begin{tabular}{lcccccc}
\toprule
\multirow{2}{*}{Methods} & \multicolumn{2}{c}{PSNR$\uparrow$} & \multicolumn{2}{c}{SSIM$\uparrow$} & \multicolumn{2}{c}{LPIPS$\downarrow$} \\
& avg & worst & avg & worst & avg & worst \\
\midrule
Random & 16.09 & 8.37 & 0.5972 & 0.2192 & 0.3400 & 0.5766 \\
FisherRF & 23.08 & 17.85 & 0.8055 & 0.6320 & 0.1063 & 0.2689 \\
AVS & 23.94 & 18.18 & 0.8479 & 0.7232 & 0.0935 & 0.2191 \\
GAUSS-MI & 16.25 & 12.71 & 0.4214 & 0.2495 & 0.3269 & 0.6968 \\
TOPIQ+GS & 18.56 & 11.12 & 0.7150 & 0.4075 & 0.2538 & 0.5031 \\
TRES+GS & 18.47 & 10.45 & 0.6803 & 0.1798 & 0.2790 & 0.6469 \\
MANIQA+GS & 14.34 & 8.08 & 0.5032 & 0.1686 & 0.4032 & 0.6348 \\
MUSIQ+GS & 15.82 & 8.64 & 0.5503 & 0.2429 & 0.3696 & 0.6419 \\
\textbf{Auto3R} & \textbf{27.00} & \textbf{21.88} & \textbf{0.8882} & \textbf{0.7819} & \textbf{0.0711} & \textbf{0.1813} \\
\bottomrule
\end{tabular}
}
\vspace{-0.1in}
\end{table}

\subsection{Evaluation of Uncertainty Quantification}

For the objects in the first row, predicted and GT images show texture differences, while the corresponding regions in GT uncertainty maps is of high-value (highlighted in red and yellow). The second row shows our predicted uncertainty and GT uncertainty. The uncertainty heatmaps between predict and GT match well, verifying the model understands reconstruction errors.

\begin{figure*}
    \centering
    \includegraphics[width=0.98\linewidth]{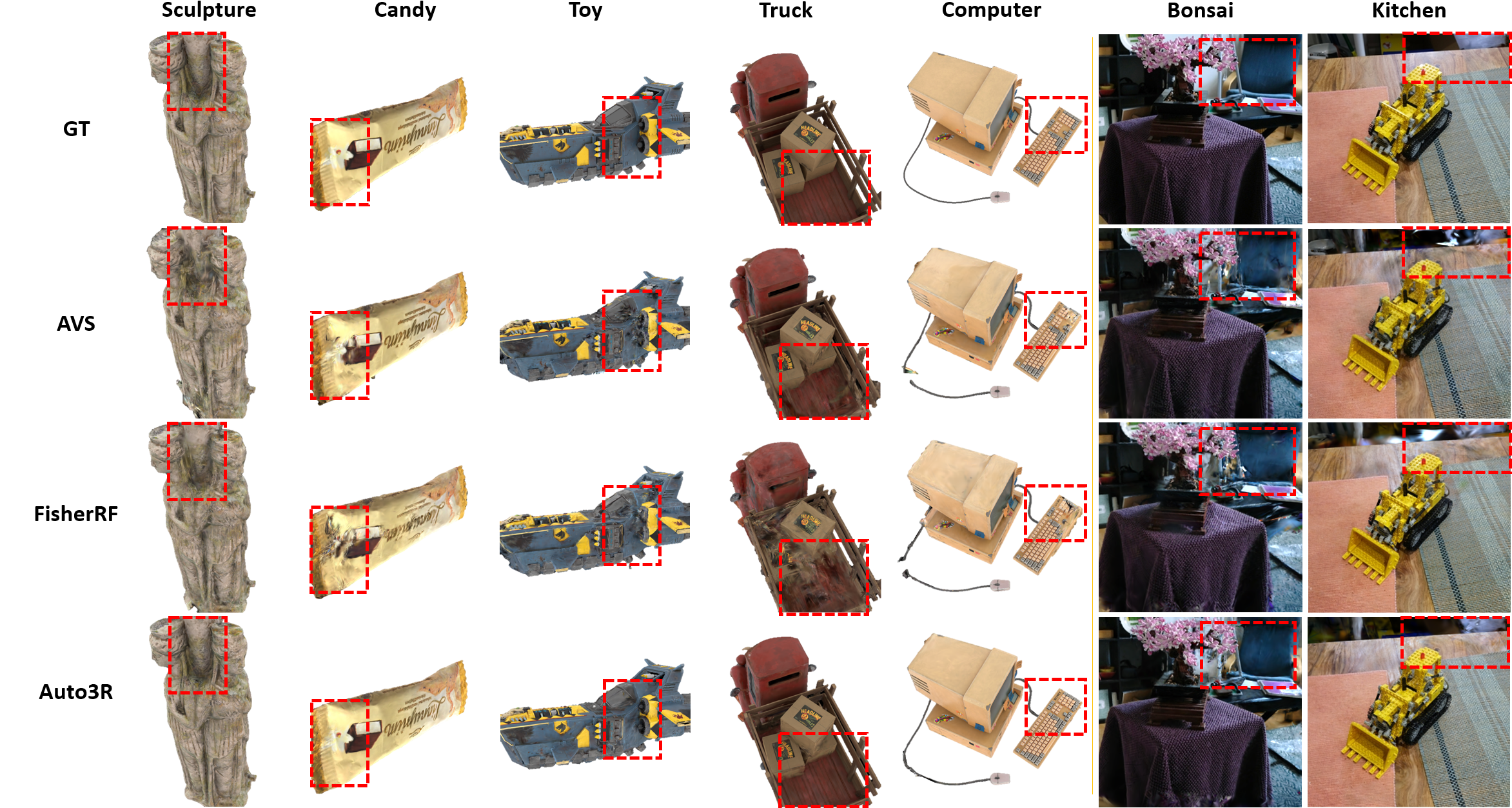}
    \vspace{-0.1in}
    \caption{\textbf{Qualitative Evaluation of Reconstructed Results}: objects (left 5 columns) and scenes (right 2 columns). Rows from top to bottom show ground truth (GT), AVS, FisherRF, and our method.}
    \label{fig:normalscenefig}
    \vspace{-0.12in}
\end{figure*}

\subsection{Evaluation of Active Object Reconstruction}
In this section, we compare our method and other state-of-the-art methods on the task of active object reconstruction. The corresponding experimental results are illustrated in Table~\ref{tab:noramaltable} and Figure~\ref{fig:normalscenefig}. %\bingqian{use proper expression for this para.}

From the quantitative results, Auto3R shows the highest PSNR and SSIM with lowest LPIPS, indicating low pixel-level error and faithful geometric structure. AVS and FisherRF also shows the acceptable results, but with some issues in the details, such as the keyboard, truck carriage in Figure~\ref{fig:normalscenefig}. For predict time, Auto3R cost about 3ms per image, with enough efficiency for real-time scanning. Generally, Auto3R achieves excellent performance in this task, outperforming the baselines in most cases.

\subsection{Evaluation of Scene Reconstruction}

To further validate the performance of our model, we carried out a larger-scale scene-level reconstruction. There are more objects and more intricate structures in the scene, which makes active reconstruction much more difficult. 

We show the quantitative results in  Table~\ref{fig:active-scene-recon}, where Auto3R delivers the best overall reconstruction quality. It performs best in PSNR and LPIPS and a close second in SSIM, preserving fine structures while suppressing artifacts. Moreover, as shown in Figure~\ref{fig:normalscenefig}, we achieve the most accurate results especially on background. These results prove our outstanding capability for large-scale scenes.

\begin{table}[t]
  \centering
  \caption{\textbf{Active scene-level reconstruction results on Mip-NeRF360}. Auto3R consistently outperforms previous uncertainty-based and image-based methods in both average and worst-case reconstruction quality.}
  \vspace{-0.1in}
  \label{fig:active-scene-recon}
    \scriptsize
    \resizebox{0.48\textwidth}{!}{
      \begin{tabular}{lccccccccc}%
        \toprule
        \multirow{2}{*}{Methods}& \multicolumn{2}{c}{PSNR$\uparrow$} & \multicolumn{2}{c}{SSIM$\uparrow$} & \multicolumn{2}{c}{LPIPS $\downarrow$} \\
        & avg & worst & avg & worst & avg & worst  \\
        \midrule
        Random &13.34&7.14&0.2341&0.1801&0.5916&0.6241  \\
        FisherRF &18.74&10.32&0.5771&0.3299&0.2619&0.4788 \\
        AVS &18.95&9.28&\textbf{0.6062}&0.3709&0.2679&0.5185  \\
        GAUSS-MI &18.22&8.99&0.5609&0.3267&0.2749&0.5142  \\
        TOPIQ+GS&13.44&7.30&0.2350 &0.1779&0.5981&0.6299  \\
        TRES+GS &13.58&6.87&0.2388 &0.1890&0.5940&0.6291  \\
        MANIQA+GS&13.46&6.64&0.2312&0.1807&0.5918&0.6338  \\
        MUSIQ+GS&13.67&6.85&0.2340 &0.1800&0.5901& 0.6137 \\
        Auto3R &\textbf{19.13}&\textbf{10.75}&0.6055&\textbf{0.3862}&\textbf{0.2520}&\textbf{0.4374}  \\
        \bottomrule
      \end{tabular}
    }
    \vspace{-0.12in}
\end{table}

\subsection{Evaluation of Specular Objects Reconstruction}
% \bingqian{we compare with other 2D UQ method, not "we implement"}
Since 3DGS is insufficient for high-quality reconstruction of specular reflective objects. We therefore compare our framework along with other 2D UQ-based methods based on GIR~\cite{GIR} (an inverse rendering 3DGS), we adjust the schedule for adding views and post-augmentation optimization parameters to accommodate its material estimation pipeline. Details of implement active reconstruction for GIR are provided in the \textbf{Supplementary Material}. And we also conduct the normal 3DGS-based framework for reconstruction, with the similar setting as above experiments.

\begin{figure}
    \centering
    \includegraphics[width=0.96\linewidth]{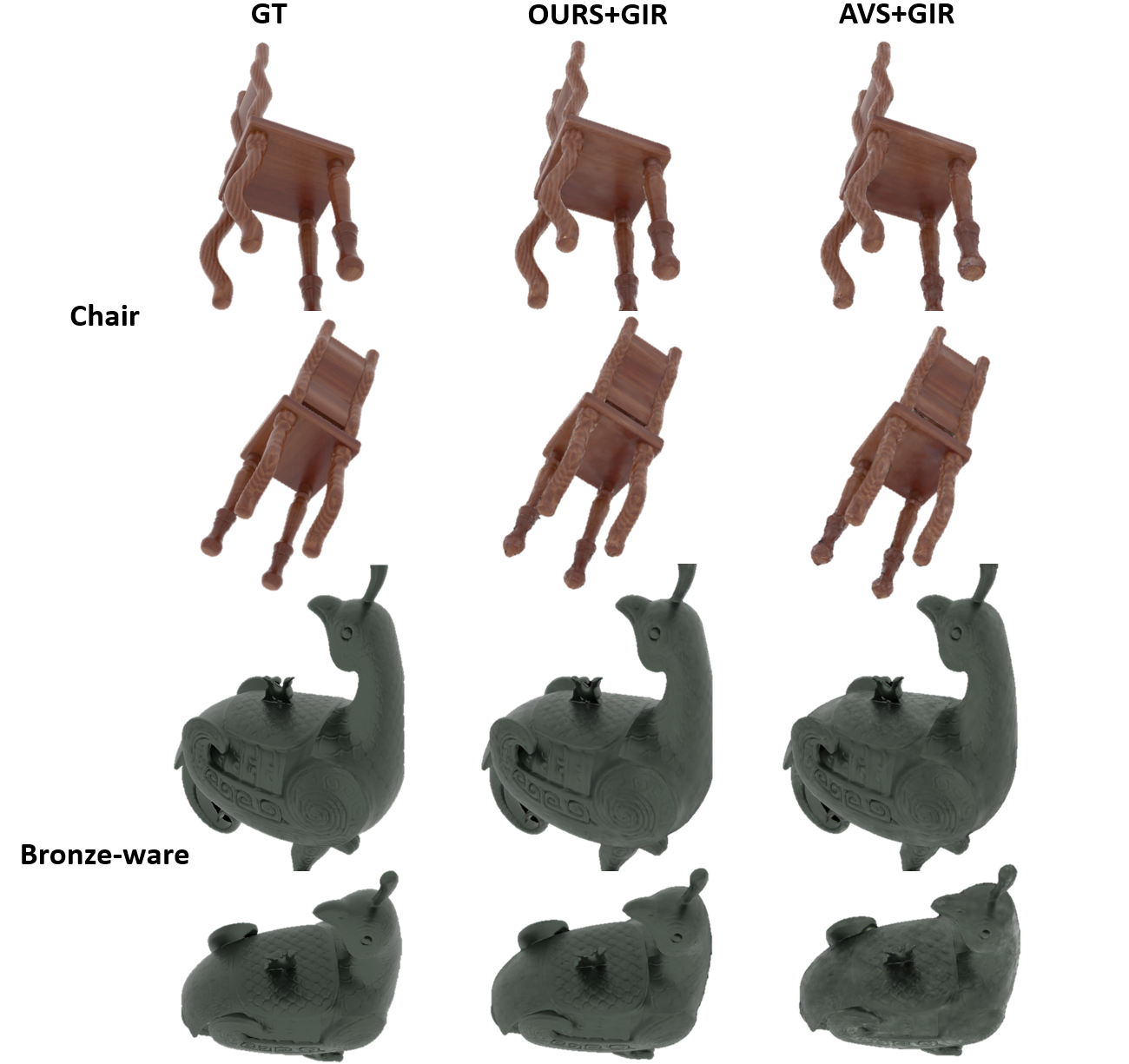}
    \vspace{-0.05in}
    \caption{Visualization of Specular Objects Reconstruction.}
    \label{fig:specularshow}
    \vspace{-0.05in}
\end{figure}

Table~\ref{tab:fanguang} and Figure~\ref{fig:specularshow} shows the quantitative results of the specular objects reconstruction, including both 3DGS-based and GIR-based methods. The results show that our Auto3R outperforms all other methods. From the figure, we can see that combining GIR with our method leads to more accurate reconstruction of regions with reflection.

\begin{SCfigure}[1][!tb]
  \centering
  \includegraphics[width=0.52\linewidth]{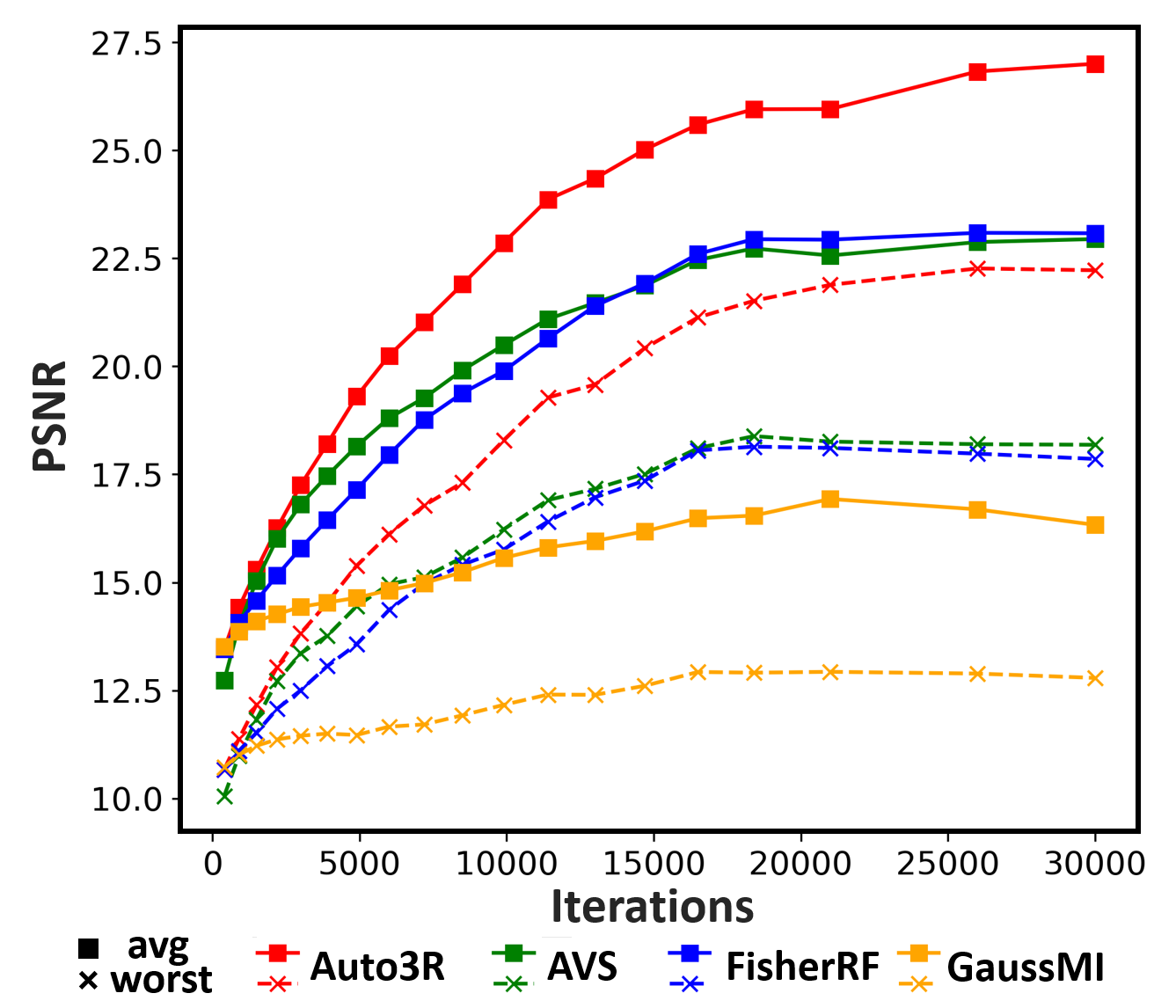}
  \caption{\textbf{Per-step curves}: average PSNR (solid) and worst-5\% PSNR (dashed) with different iterations of next view sampling for each method. Colors differentiate methods.}
  \label{fig:active-curves-left}
  \vspace{-0.05in}
\end{SCfigure}

\begin{table}[t]
\vspace{-0.03in}
\centering
\caption{\textbf{Active reconstruction results for specular objects.} We evaluate both 3DGS-based and GIR-based frameworks. Auto3R significantly improves reflection fidelity and quantitative accuracy under challenging non-Lambertian surfaces.}
\vspace{-0.1in}
\label{tab:fanguang}
\scriptsize
\resizebox{0.48\textwidth}{!}{
\begin{tabular}{llcccc}
\toprule
\multirow{2}{*}{Framework}&\multirow{2}{*}{Methods}& \multicolumn{2}{c}{PSNR$\uparrow$} & \multicolumn{2}{c}{SSIM$\uparrow$} \\
& & avg & worst & avg & worst \\
\midrule
\multirow{6}{*}{3DGS}
& Random & 17.69 & 12.51 & 0.6483 & 0.3563 \\
& FisherRF & 22.84 & 17.66 & 0.8248 & 0.6715 \\
& AVS & 23.82 & 18.61 & 0.8652 & 0.7401 \\
& TOPIQ+GS & 20.92 & 14.69 & 0.7719 & 0.4593 \\
%& TRES+GS & 20.03 & 14.50 & 0.7162 & 0.4328 \\
& \textbf{Auto3R} & \textbf{26.41} & \textbf{20.94} & \textbf{0.8904} & \textbf{0.7769} \\
\midrule
\multirow{3}{*}{GIR}
& Random & 26.85 & 20.09 & 0.8855 & 0.8278 \\
& AVS & 29.53 & 20.36 & 0.9417 & 0.8319 \\
& \textbf{Auto3R} & \textbf{30.29} & \textbf{22.92} & \textbf{0.9505} & \textbf{0.8791} \\
\bottomrule
\end{tabular}
}
\vspace{-0.2in}
\end{table}

\vspace{-0.02in}
\subsection{Ablation Study}
\vspace{-0.02in}
We conduct a detailed ablation study to evaluate the contribution of two key components in our uncertainty quantification: (1) the \textbf{depth-aware blending}, which aggregates uncertainty along depth to better reflect 3D coverage, and (2) the \textbf{depth-uncertainty reweighting}, which adaptively adjusts the blending weights based on confidence in depth estimation. The results are summarized in Table~\ref{tab:ablation}.

\textbf{Effect of Depth Blending.}
Without depth blending (only relies on 2D appearance cues), the model ignores geometric consistency across viewpoints. This leads to unstable uncertainty estimation, particularly in occluded or depth-varying regions. Introducing depth blending improves PSNR from 22.48 to 23.41 (+0.93) and SSIM from 0.7818 to 0.8124 (+0.0306), showing that incorporating geometric cues provides a reliable indicator of poorly reconstructed areas. Among different formulations, the linear depth weighting performs slightly better than quadratic weighting, suggesting that moderate depth dependence balances local and global uncertainty aggregation effectively.

\textbf{Effect of Depth-Uncertainty Reweighting.}
Further adding depth-uncertainty reweighting consistently enhances reconstruction quality across all metrics. By dynamically suppressing unreliable depth regions in early reconstruction stages, this mechanism refines the uncertainty field and prevents overconfident sampling near ambiguous geometry. With this module, PSNR rises from 23.41 to 23.77 and SSIM from 0.8124 to 0.8252, confirming that coupling uncertainty prediction with self-assessed depth confidence improves both accuracy and stability of view selection.

\textbf{Overall Analysis.}
Combining both depth-aware blending and uncertainty reweighting yields the best overall performance, accelerating convergence toward high-fidelity reconstructions. The ablation validates that each component contributes complementary benefits—depth blending introduces geometric awareness, while reweighting stabilizes uncertainty estimation throughout iterative reconstruction.

\begin{table}[t]
\centering
\caption{Ablation on depth blending and depth-uncertainty reweighting. Both contribute to consistent gains in PSNR and SSIM, validating the effectiveness of our uncertainty formulation.}
\vspace{-0.08in}
\tiny
\resizebox{0.48\textwidth}{!}{
\begin{tabular}
{lcccc}
    \toprule
    \multirow{2}{*}{Ablation} & \multicolumn{2}{c}{PSNR$\uparrow$} &\multicolumn{2}{c}{SSIM$\uparrow$} \\

     & avg & worst&avg &worst\\
    \midrule
    w/o depth-uq &23.41&17.95&0.8124&0.6494\\
    w/o depth-blending &22.60&15.94&0.7910&0.5664\\
    $\text{depth}^2$-blending&22.59&17.60&0.8131&0.6388\\
ours&\textbf{23.77}&\textbf{18.49}&\textbf{0.8252}&\textbf{0.6703}\\
    \bottomrule
   
\end{tabular}
% \bingqian{split the two factor ablation in table}

\label{tab:ablation} 
}
\end{table}
\vspace{-0.05in}
\section{Real-world Robotic Deployment}
\begin{figure}
    \centering
    \includegraphics[width=0.94\linewidth]{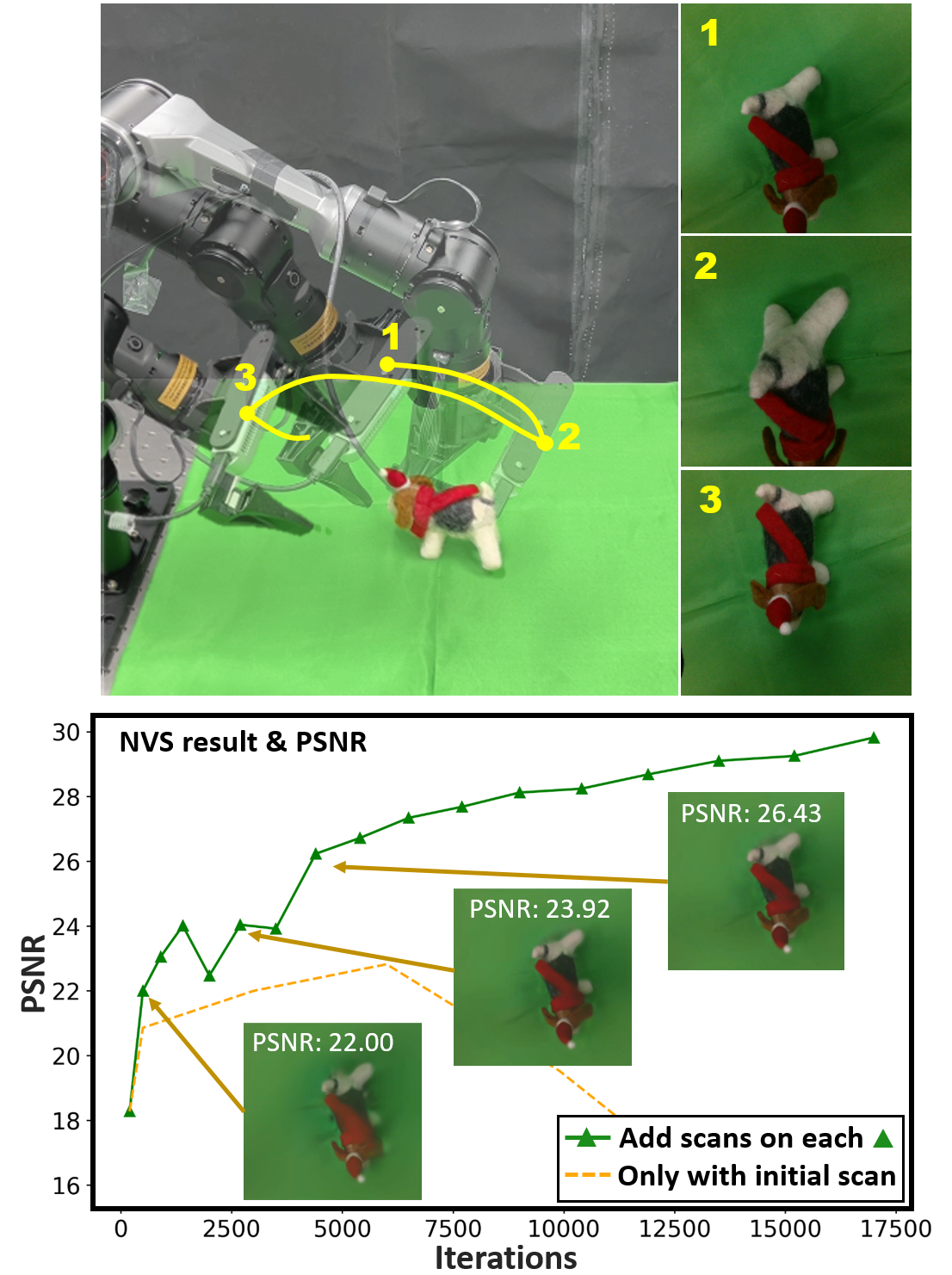}
    \vspace{-0.05in}
    \caption{Top: the first three scanning path of the robot. Bottom: the novel view rendered images and PSNR metrics by Auto3R-extension after each scan.}
    % \vspace{-0.1in}
    \label{fig:robot}
\end{figure}
We apply our method on a real-world robot to scanning various objects, a toy made of fabric, and a circuit board. We use the agilex piper robot, with an Intel RealSense D435 fixed on the robot end. The robot has a reach of 626 mm and its end-effector supports 6-DOF adjustments. Images are captured at a resolution of 640 × 480. 

The layout and the scanning process of the robot is shown in the Figure~\ref{fig:robot}. And the scanning process with the real-time novel view rendering performance can be find in the video in \textbf{Supplementary Material}.

% \begin{table}[]
%     \centering
%     \caption{The quantitative results on robotic scanning task. We compare our Auto3R with AVS and FisherRF in PSNR and SSIM on novel view.}
%     \vspace{-0.1in}
%     \resizebox{0.3\textwidth}{!}{
%     \begin{tabular}{lcc}
%     \toprule
%         Methods&PSNR$\uparrow$&SSIM$\uparrow$\\
%          \midrule
%          Auto3R-extension&28.56&0.8953 \\
%          Auto3R&22.91&0.8346 \\
%          AVS&20.35&0.8159 \\
%          FisherRF&18.95&0.7415 \\
%          \bottomrule
%     \end{tabular}
%     }
%     \vspace{-0.05in}
    
%     \vspace{-0.25in}
%     \label{tab:robot}
% \end{table}

\begin{SCtable}[\sidecaptionrelwidth][t]  
  \caption{Reconstruction quality under \textbf{scanning with Robots}.}
  \label{tab:robot}
  \footnotesize
  \setlength{\tabcolsep}{8pt}   
  \begin{tabular}{lcc}
    \toprule
    Methods & PSNR$\uparrow$ & SSIM$\uparrow$ \\
    \midrule
    Auto3R-extension & \textbf{28.56} & \textbf{0.8953} \\
    Auto3R           & 22.91 & 0.8346 \\
    AVS              & 20.35 & 0.8159 \\
    FisherRF         & 18.95 & 0.7415 \\
    \bottomrule
  \end{tabular}
  \vspace{-0.1in}
\end{SCtable}

 We also applied Auto3R and its extension (described in Sec.~\ref{sec:4.3}), AVS and FisherRF on robotic scanning task. For each task, we start with 4 views in a forward path. For extension of Auto3R, we predict a scanning path and scan an image sequence on the path, for remainning baselines, we keep the same intervals for sampling new view. Table~\ref{tab:robot} shows the results of each method, where the extension of Auto3R achieve the most accurate results, indicating that the effectiveness of extension UQ for scanning path. The remaining results indicate that that our Auto3R outperforms the other baselines on these real-world  reconstruction case.

\vspace{-0.05in}
\section{Conclusion}
\vspace{-0.05in}
\label{sec:conclusion}
In this work, we presented \textbf{Auto3R}, a fully automated framework for active 3D reconstruction and scanning. By introducing a data-driven uncertainty quantification model that jointly reasons over rendered color and depth, Auto3R enables accurate, geometry-aware, and material-robust view planning without manual intervention. Extensive experiments demonstrate that Auto3R achieves superior reconstruction quality compared to existing uncertainty or information based approaches, and can be deployed on robotic systems for real-world digitization tasks.

% In future work, we plan to integrate a feed-forward 3DGS architecture with our uncertainty model to enable low-latency prediction and on-the-fly active scanning, paving the way toward real-time, autonomous, and scalable 3D scene understanding.
{
    \small
    \bibliographystyle{plain}
    \bibliography{main}
}
\appendix

\clearpage
%\setcounter{page}{1}
%\maketitlesupplementary
\section*{Supplementary Materials}
\label{sec:supp}

\section{Implementation Details on GIR}

\noindent\textbf{Training procedure.} 
Unlike 3DGS, the optimization of GIR is organized into three consecutive stages. The first stage performs coarse geometry fitting similar to standard 3DGS initialization. This stage runs for 25,000 iterations, enabling directional masking and refining Gaussian surface normals. The second stage lasts for 30,000 iterations and extends optimization to material and illumination parameters. In our active reconstruction setting, when a new view is introduced, we first re-optimize the geometry before updating appearance-related parameters. This step is essential to prevent the network from locking onto incorrect geometry caused by premature material fitting. Therefore, directional masking is temporarily disabled for a few iterations after each new view is added.

\begin{figure}[h]
    \centering
    \includegraphics[width=\linewidth]{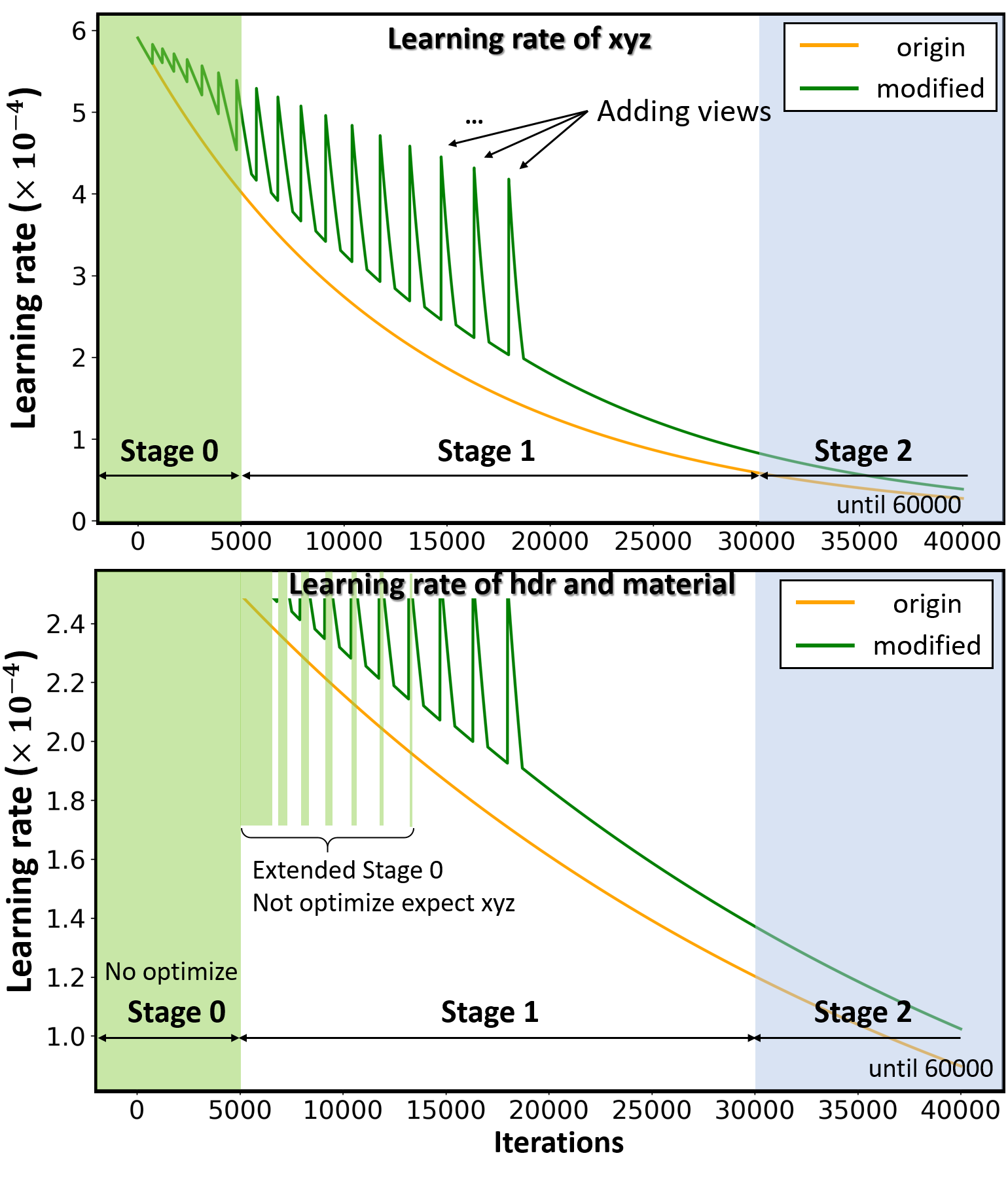}
    \caption{Learning rate schedule of original GIR (yellow) and our Auto3R-adapted GIR (green). The top and bottom plots correspond to the learning rates for geometry and for appearance-related parameters (materials, illumination).}
    \label{fig:supplr}
\end{figure}

\noindent\textbf{Learning rate schedule.} 
The original GIR learning rate schedule is shown by the yellow curve in Fig.~\ref{fig:supplr}. The learning rates for the Gaussian centers ($xyz$) and for material and lighting parameters decay at the beginning of the first and second stages, respectively. We denote the learning rate function as
\begin{equation}
lr = s(\textit{iter}),
\end{equation}
where $\textit{iter}$ indicates the current iteration number.  
In our implementation, after each new view is added, we reset the learning rate to a higher value to allow rapid geometric convergence, followed by a short exponential decay for $\textit{iter}_d = 720$ iterations:
\begin{equation}
lr_d = \frac{base}{4} + \frac{(\textit{iter} - base)(base/2 + \textit{iter}_d)}{\textit{iter}_d},
\end{equation}
where \textit{base} denotes the iteration index of the most recent view addition. After this rapid adjustment, we resume optimization with a standard learning rate schedule, 
\begin{equation}
lr_c = s(\textit{iter} - base/4),
\end{equation}
until the next new view is incorporated.

\section{More Results on Robots}

\noindent\textbf{About the Video.} 
The supplementary video contains three synchronized visual streams. The \textbf{left} panel shows a top-down view of the scene, capturing the robot’s real-time motion. The \textbf{top-right} panel displays the live RGB feed from the robot-mounted camera, which also serves as input for reconstruction. The \textbf{bottom-right} panel shows periodically updated renderings of the current reconstruction from multiple test viewpoints as the optimization progresses.

\begin{figure*}[t]
    \centering
    \includegraphics[width=0.9\linewidth]{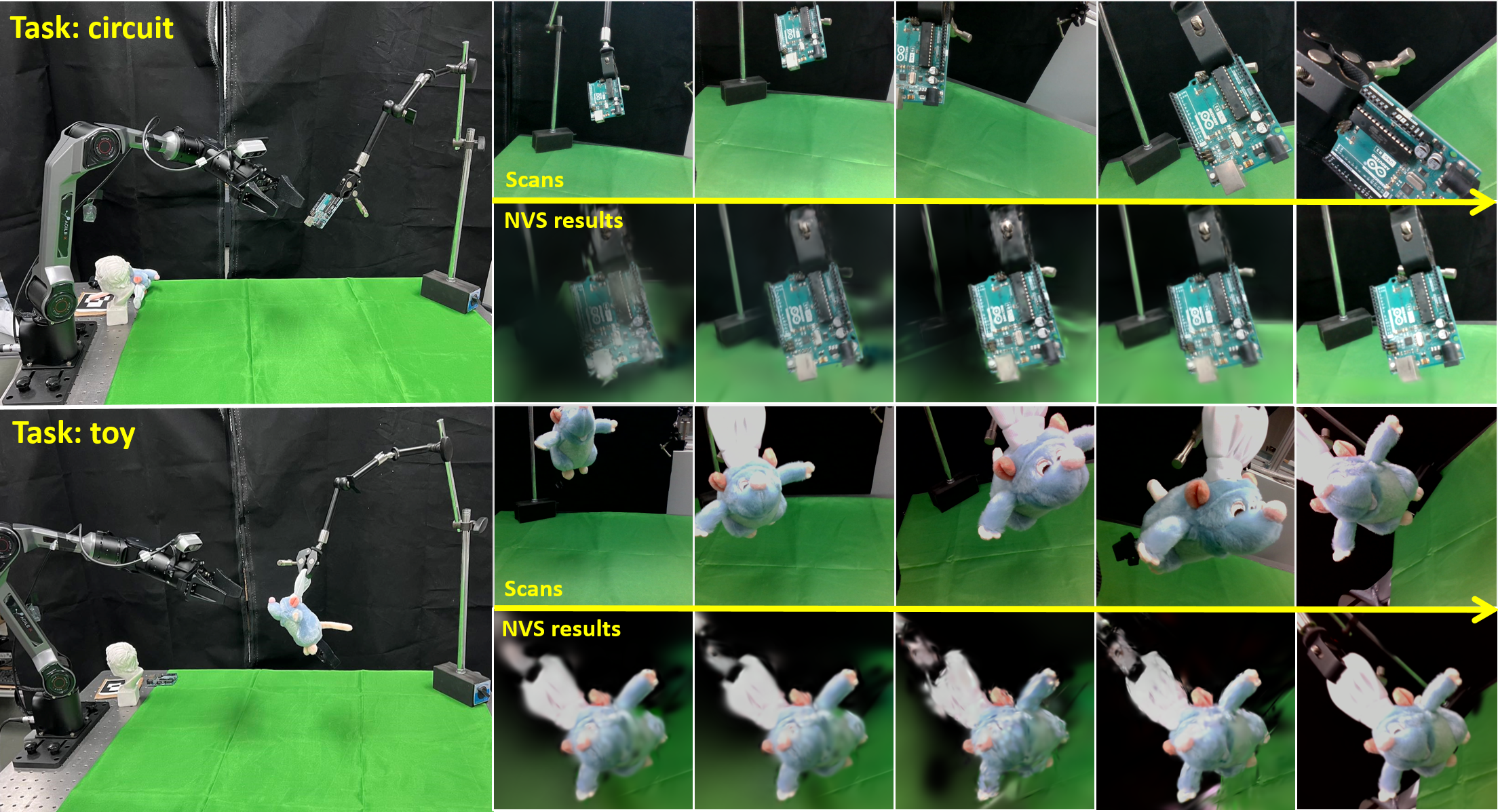}
    \caption{
    The scanning layout, captured views, and reconstruction results. 
    Each row corresponds to one robotic scanning task (circuit board and toy). 
    For each object, the left shows the layout and robot trajectory, while the right shows captured images followed by reconstructed novel views.}
    \label{fig:supprobo}
\end{figure*}

\noindent\textbf{Reconstruction Results.} 
We provide additional qualitative results of robotic object scanning and reconstruction. Figure~\ref{fig:supprobo} presents two representative examples: a circuit board and a plush toy. Each example includes the scanning layout, several images captured by the robot during motion, and reconstructed renderings from multiple viewing angles. The results demonstrate that Auto3R produces accurate and stable reconstructions in real robotic scenarios, highlighting its potential for embodied active perception and automated digitization tasks.

\section{Detailed Results for Experiments}

\begin{figure*}[t]
    \centering
    \includegraphics[width=\linewidth]{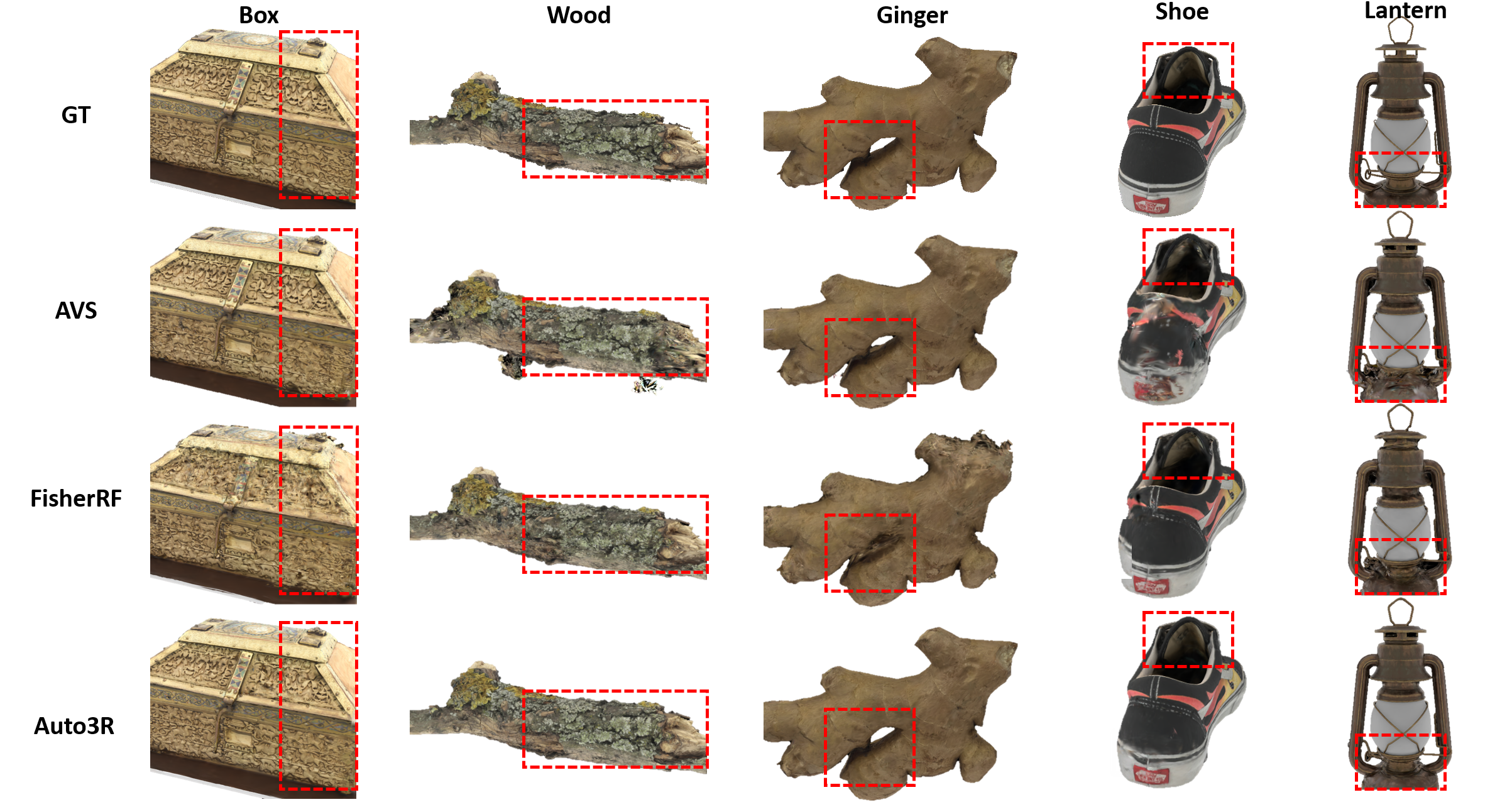}
    \caption{Additional qualitative reconstruction results on five objects.  
    From top to bottom: ground truth (GT), AVS, FisherRF, and our Auto3R.}
    \label{fig:suppobj}
\end{figure*}

\begin{figure*}[t]
    \centering
    \includegraphics[width=\linewidth]{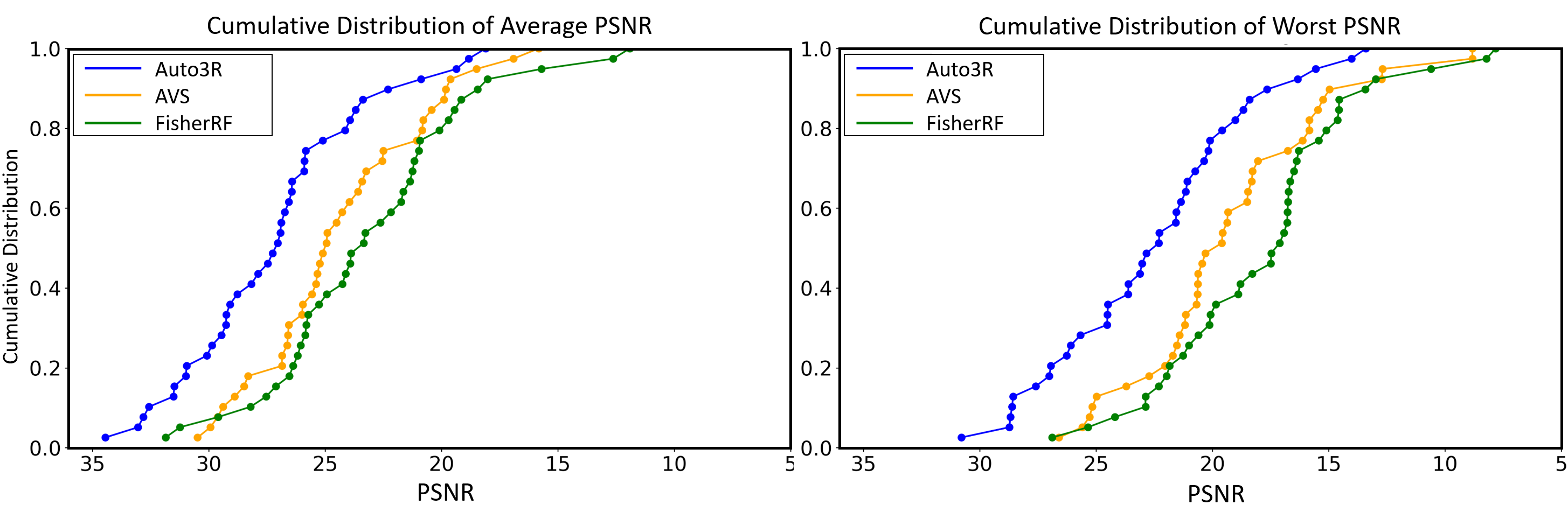}
    \caption{Cumulative distribution of the average PSNR (left) and worst-case PSNR (right) across all test objects.  
    Blue, yellow, and green curves represent Auto3R, AVS, and FisherRF, respectively.}
    \label{fig:suppcdf}
\end{figure*}

\noindent\textbf{Per-object results.} 
We provide additional details for the automated object reconstruction experiments.  
Figure~\ref{fig:suppcdf} presents the cumulative distribution of both the average and worst-case PSNR across all test objects, offering a more comprehensive view of the performance differences between methods.  
Figure~\ref{fig:suppobj} shows qualitative comparisons on additional objects, including ground-truth renderings and reconstructed results from AVS, FisherRF, and our Auto3R.

\noindent\textbf{Per-scene results.}

\begin{table*}

\caption{Per scene NVS results on Mip-NeRF360 dataset}
\label{tab:suppscene}
\centering
  \begin{tabular*}{0.93\textwidth}{ccccccccccccc}
    \toprule
    \multicolumn{1}{c}{Scene} & \multicolumn{2}{c}{Auto3R} &\multicolumn{2}{c}{AVS} & \multicolumn{2}{c}{fisherrf} & \multicolumn{2}{c}{GaussMI} & \multicolumn{2}{c}{TOPIQ+GS}& \multicolumn{2}{c}{TRES+GS} \\
    & Avg & Worst & Avg & Worst & Avg & Worst &  Avg & Worst &  Avg & Worst&  Avg & Worst \\
    \midrule
    Bicycle &\textbf{15.07}&2.93&14.79&2.70&14.48&3.00&13.97&2.60&13.31&5.93&12.72&4.09\\
    Bonsai&\textbf{22.45}&14.89&21.47&10.78&22.06&12.43&21.36&11.78&13.11&7.40&13.19&9.11\\
    Counter&21.56&13.82&21.45&12.97&\textbf{21.78}&14.99&21.30&12.50&13.11&9.41&13.27&8.05\\
    Garden&\textbf{21.99}&16.98&21.84&15.04&21.14&13.53&20.96&13.24&13.44&10.48&15.08&9.84\\ 
    Kitchen&22.73&13.63&22.45&12.29&\textbf{23.11}&14.28&22.02&13.52&14.11&10.06&14.26&10.31\\ 
    Room&24.42&14.54&\textbf{24.77}&14.85&23.69&14.50&23.09&11.83&11.93&5.56&12.18&6.69\\ 
    Stump&\textbf{18.45}&9.52&18.38&7.83&18.13&9.77&17.80&7.01&16.84&7.60&17.03&5.97\\
    Flowers&12.57&4.49&\textbf{12.68}&4.23&11.70&4.42&11.98&3.92&12.53&4.24&12.17&4.07\\
    Treehill&\textbf{12.90}&5.99&12.78&5.51&12.70&5.98&11.57&4.53&12.51&5.00&12.32&3.71\\
    \hline
    average&\textbf{19.13}&10.75&18.95&9.28&18.74&10.32&18.22&8.99&13.43&7.30&13.58&6.87\\
    \bottomrule
  \end{tabular*}
\end{table*}

We further report scene-level reconstruction results on the Mip-NeRF360 dataset.  
Due to space limitations, Table~\ref{tab:suppscene} summarizes the results for the six best-performing methods.  
The table lists the average and worst-case PSNR for each scene, providing a detailed view of method performance across diverse environments.

\section{Future Work}

In future work, we plan to further integrate our framework into real-world robotic applications.  
On the systems side, we aim to incorporate path-length optimization, autonomous obstacle avoidance, and motion-level planning into the active reconstruction pipeline.  
These enhancements will enable the robot to navigate more efficiently and safely while selecting informative viewpoints, ultimately improving both the reliability and scalability of automated 3D reconstruction in practical environments.

\end{document}